\documentclass[conference]{IEEEtran}
\usepackage[latin9]{inputenc}
\usepackage{times}
\usepackage{epsfig}
\usepackage{graphicx}
\usepackage{amsmath}
\usepackage{amssymb}
\usepackage{multirow}
\usepackage{amsthm}
\usepackage{subfigure}

\newcommand{\etal}{\textit{et al.}}

\usepackage[pagebackref=true,breaklinks=true,letterpaper=true,colorlinks,bookmarks=false]{hyperref}

\begin{document}
%
\title{Detecting and Simulating Artifacts in GAN Fake Images (Extended Version)}

\author{
\IEEEauthorblockN{Xu Zhang, Svebor Karaman, and Shih-Fu Chang}
\IEEEauthorblockA{Columbia University, 
Email:  \{xu.zhang,svebor.karaman,sc250\}@columbia.edu}}



\maketitle

\vspace{-0.5em}
\begin{abstract}
To detect GAN generated images, conventional supervised machine learning algorithms require collection of a number of real and fake images from the targeted GAN model. 
However, the specific model used by the attacker is often unavailable. 
To address this, we propose a GAN simulator, AutoGAN, which can simulate the artifacts produced by the common pipeline shared by several popular GAN models. 
Additionally, we identify a unique artifact caused by the up-sampling component included in the common GAN pipeline. 
We show theoretically such artifacts are manifested as replications of spectra in the frequency domain and thus propose a classifier model based on the spectrum input, rather than the pixel input. 
By using the simulated images to train a spectrum based classifier, even without seeing the fake images produced by the targeted GAN model during training, our approach achieves state-of-the-art performances on detecting fake images generated by popular GAN models such as CycleGAN. 
\end{abstract}

\IEEEpeerreviewmaketitle

\section{Introduction}
\label{sec:intro}
%

Machine learning based approaches, such as those based on Generative Adversarial Network (GAN)~\cite{goodfellow2014generative}, have made creation of near realistic fake images much more feasible than before and have enabled many interesting applications in entertainment and education. 
Some high-resolution images generated by the latest GAN models are hardly distinguishable from real ones for human viewers~\cite{karras_style-based_2018,brock_large_2018}. 
However, this also raises concerns in security and ethics as the traditional perspective of treating visual media as trustworthy content is not longer valid.
As a partial remedy, the development of an automatic tool to distinguish real from GAN generated images will provide great value. 

%
%
%
%

A typical way to design a real vs. GAN fake image classifier is to collect a large number of GAN generated images from one or multiple pre-trained GAN models and train a binary classifier~\cite{marra_detection_2018,yu_attributing_2018}. 
Unfortunately, in real world applications, we generally have no access to the specific model used by the attacker. 
To train a classifier with fewer or even no fake image from the pre-trained GAN model, we explore two directions. 
1) We identify the key up-sampling component of the generation pipeline and theoretically show the unique artifacts generated by this component in the frequency domain, thus calling for the use of spectrum, rather than pixels, as input to GAN image classifiers.
2) We develop an emulator framework which simulates the common generation pipeline shared by a large class of popular GAN models. 

First, we study the GAN generation pipeline and find out that many popular GAN models such as CycleGAN~\cite{zhu_unpaired_2017} and StarGAN~\cite{choi_stargan:_2017} share common up-sampling layers. 
The up-sampling layer is one of the most important modules of GAN models, as it produces high-resolution image/feature tensors from low-resolution image/feature tensors. 
Odena~\etal~\cite{odena_deconvolution_2016} show that the up-sampling layer using transposed convolution leaves checkerboard artifacts in the generated image. 
In this paper, we extend the analysis in the frequency domain and use signal processing properties to show that up-sampling results in replications of spectra in the frequency domain.
To directly discriminate the GAN induced up-sampling artifacts, we propose to train a classifier using the frequency spectrum as input instead of the raw RGB pixel. We will show through experiments 
that the spectrum-based classifier trained even with images from only one semantic category (e.g. real and fake horse images) generalizes well to other unseen categories. 

We further propose to address the situation where there is no access to pre-trained GAN models by using a GAN simulator, AutoGAN. 
Using only real images during training, AutoGAN simulates the GAN generation pipeline and generates simulated ``fake'' images. 
Then the simulated images can be used in classifier training.
%
Experiment shows that, although having never seen any fake image generated by CycleGAN, the model trained with simulated images still achieves the  state-of-the-art performance on CycleGAN data. 
It outperforms all methods that require access to the fake images generated by the actual GAN models used in image faking.
%

In summary, the paper makes the following contributions:
\begin{enumerate}
    \item This is the first work proposing a novel approach based on the GAN simulator concept to emulate the process commonly shared by popular GAN models. Such a simulator approach frees developers of the requirement of having access to the actual GAN models used in generating fake images when training the classifier.
    \item We revisit the artifact inducted by the up-sampling module of GANs and present a new signal processing analysis, from which we propose a new approach to the classifier design based on the spectrum input.
\end{enumerate}

\section{Related Work}
\label{sec:related_work}
%
In image generation, GAN~\cite{goodfellow2014generative} can be applied in the following scenarios: 
1) taking noise as input to produce an image~\cite{karras_progressive_2017,karras_style-based_2018,brock_large_2018}; 
2) taking an image from one semantic category (such as horse) as input to produce an image of another semantic category (such as zebra)~\cite{zhu_unpaired_2017,choi_stargan:_2017,park_semantic_2019}; 
3) taking a sketch or a pixel level semantic map as input to produce a realistic image that is constrained by the layout of the sketch~\cite{isola2017image, zhu_unpaired_2017,park_semantic_2019}. 
The latter two scenarios give users more control to the generated content since a specific input is chosen and the output is expected to respect properties of that input. 

In response, the forensics community has been working on detecting such generated content~\cite{marra_detection_2018,li_ictu_2018,li_exposing_2018,agarwal_protecting_nodate}. 
Marra~\etal~\cite{marra_detection_2018} propose to use raw pixels and conventional forensics features extracted from real and fake images to train a classifier.
%
Nataraj~\etal~\cite{nataraj_detecting_2019} propose to use the co-occurrence matrix as the feature and show better performance than that of classifiers trained over raw pixels on CycleGAN data. 
McCloskey and Albright~\cite{mccloskey_detecting_2018} observe that GAN generated images have some artifacts in color cues due to the normalization layers. These artifacts can be exploited for detection. 
\cite{marra2019gans} and \cite{albright2019source} study the fingerprints
of GAN models.
All the machine learning based methods require sufficient training images generated by one or multiple pre-trained GAN models to ensure the generalization ability of the classifier.
%

In real-world applications, it is often not possible to have access to the pre-trained GAN model used in generating the fake image. 
We study how to remove such requirements of accessing pre-trained models when training GAN fake image classifier by understanding, detecting and simulating the artifact induced in the GAN generation pipeline.

\section{Up-sampling Artifacts in GAN Pipelines}
\label{sec:revisit}
\subsection{GAN Pipelines}
\label{sec:pipeline}
\begin{figure}[tbp]
\begin{center}
	\includegraphics[width=0.9\linewidth]{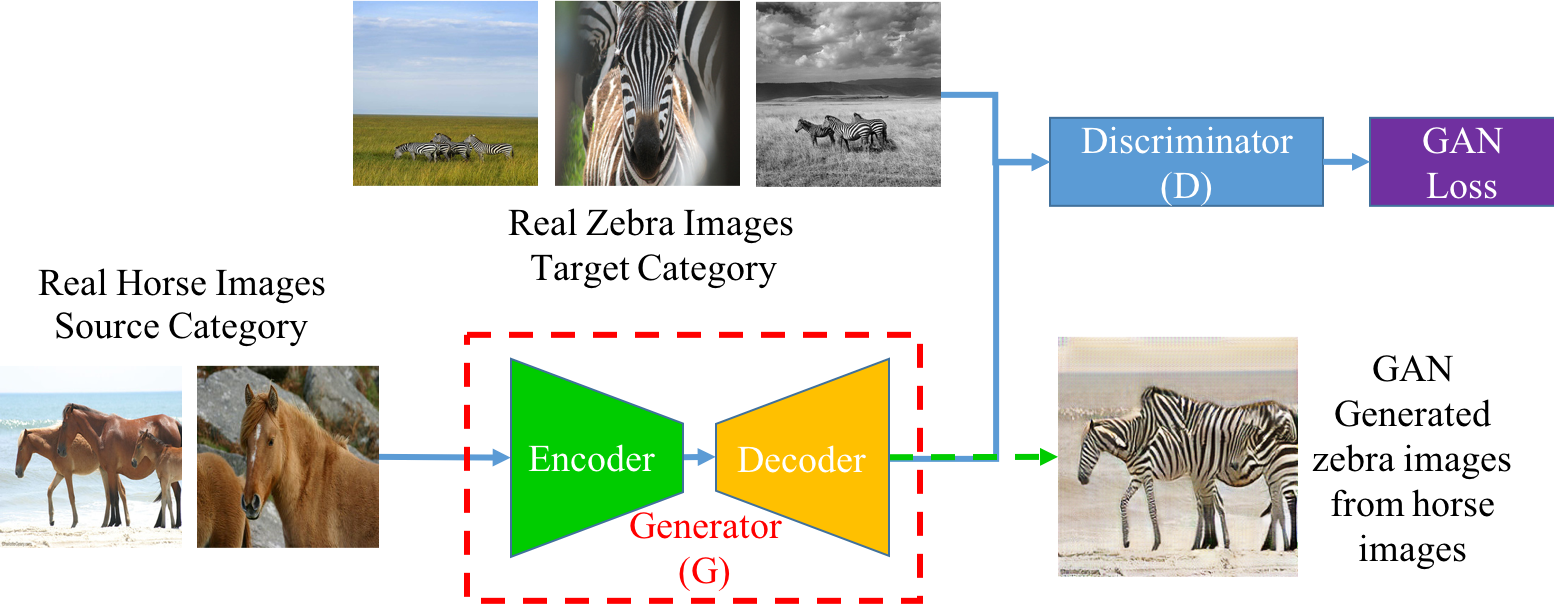}
\end{center}
\vspace{-1.0em}
\caption{Typical pipeline for image2image translation.}
\label{fig:pipeline}
\vspace{-1.0em}
\end{figure}

We first review the general pipeline for image2image or sketch2image translation, as illustrated in 
Fig.~\ref{fig:pipeline}. 
During the training phase, an image translation model takes images from two categories (e.g. horse/zebra) as input and learns to transfer images from one category (source) to the other (target). 
It contains two main components: discriminator and  generator. 
The discriminator tries to distinguish real images of the target category from those generated by the generator. 
The generator takes an image of the source category as input and tries to generate an image that is similar to images of the target category, making them indistinguishable by the discriminator. 
The generator and the discriminator are alternatively trained until reaching an equilibrium.
During the generation phase, an image from the source category is passed through the generator to obtain an image similar to the target category. 

We show more details of the generator, since it directly synthesizes the fake image. 
As shown in Fig.~\ref{fig:pipeline}, the generator contains two components, encoder and decoder. 
The encoder contains a few down-sampling layers which try to extract high-level information from the input image and generate a low-resolution feature tensor. The decoder, on the other hand, contains a few up-sampling layers which take the low-resolution feature tensor as input and output a high-resolution image. 
It is important to understand how the decoder renders fine details of the final output image from the low-resolution feature tensor with the up-sampler.

\subsection{The Up-sampler}
\label{sec:upsampler}
Although the structures of GAN models are quite diverse, the up-sampling modules used in different GAN models are consistent.
Two most commonly used up-sampling modules in the literature are transposed convolution (a.k.a  deconvolution)~\cite{radford_unsupervised_2015,zhu_unpaired_2017,choi_stargan:_2017} and nearest neighbor interpolation~\cite{karras_progressive_2017,park_semantic_2019}. 
\begin{figure}[tb]
\vspace{-1.0em}
\begin{center}
	\includegraphics[width=0.8\linewidth]{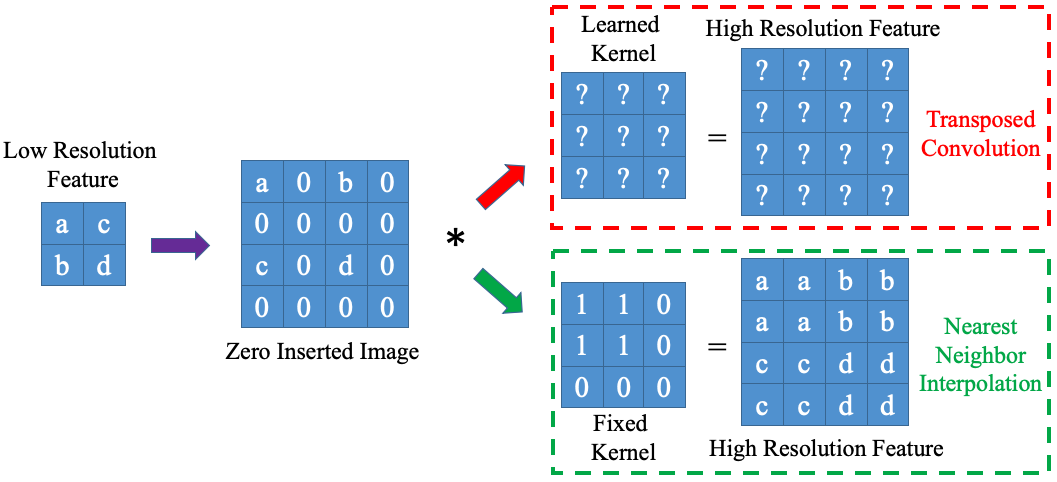}
\end{center}
\vspace{-1.0em}
\caption{Transposed convolution and nearest neighbor interpolation.}
\label{fig:upsampling}
\vspace{-1.0em}
\end{figure}
\begin{figure}[b]
\vspace{-1.0em}
\begin{center}
	\includegraphics[width=0.9\linewidth]{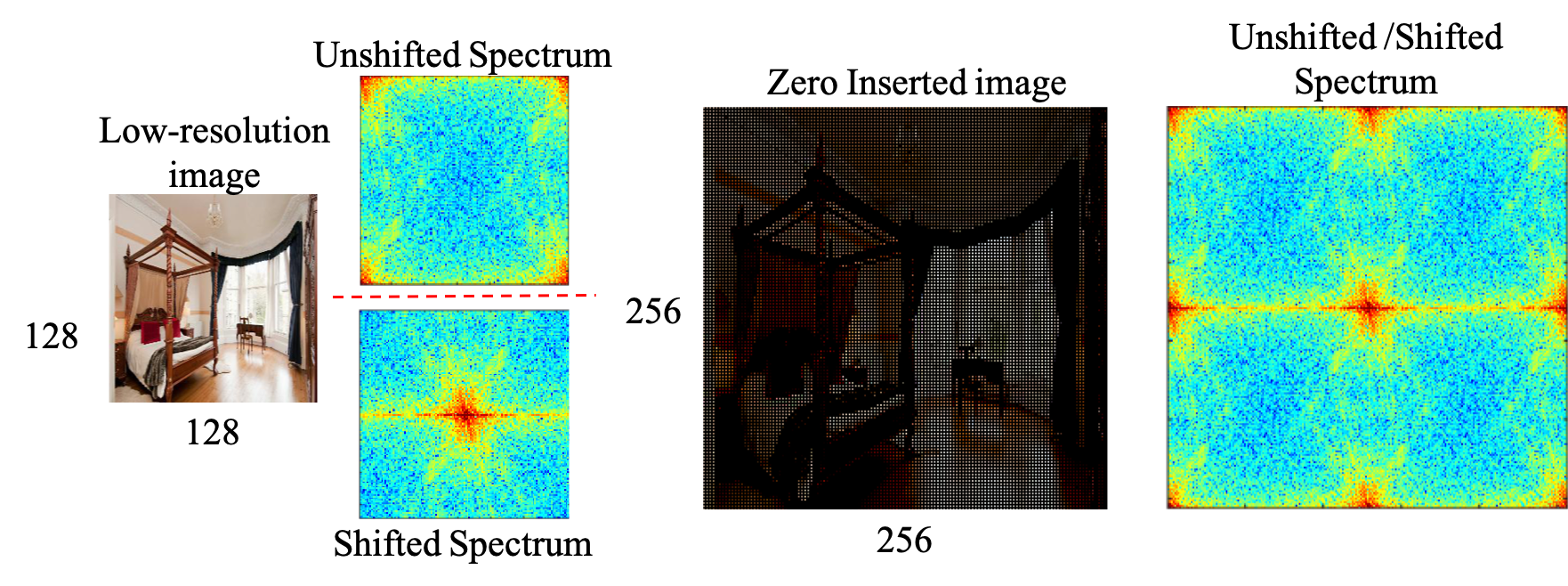}
\end{center}
\vspace{-1.5em}
\caption{The spectrum of the low resolution image and the spectrum of the zero inserted image.}
\label{fig:spectrum}
\vspace{-1.0em}
\end{figure}
Interestingly, both up-samplers can be formulated as a simple pipeline, shown in Fig.~\ref{fig:upsampling}, where $\star$ denotes convolution. 
Given a low-resolution feature tensor as input, the up-sampler increases both the horizontal and vertical resolutions by a factor of $m$. 
For illustration purposes, we assume $m=2$, which is the most common setting. 
The up-sampler inserts one zero row/column after each row/column in the low-resolution feature tensor and applies a convolution operation in order to assign appropriate values to the ``zero-inserted'' locations. 
The difference between transposed convolution and nearest neighbor interpolation is that the convolution kernel in transposed convolution is learnable, while in the nearest neighbor interpolation it is fixed (as shown in Fig.~\ref{fig:upsampling}). 

The up-sampling artifact of transposed convolution is called ``checkerboard artifact'' and has been studied by Odena~\etal~\cite{odena_deconvolution_2016} in the spatial domain. 
Here we provide further analysis in the frequency domain.
According to the property of Discrete Fourier Transform~(DFT), as shown in Fig.~\ref{fig:spectrum}, inserting zero to the low-resolution image is equivalent to replicating multiple copies of the spectrum of the original low-resolution image over the high frequency part of the spectrum of the final high-resolution image. For illustration purpose, we show the spectrum of the gray-scale image. Warmer color means higher value.

We prove it in the 1D case and it can be easily extended to the 2D case for images.  
Assume the low-resolution signal $x(n), n = 0,\ldots,N-1$ has $N$ points and its DFT is $X(k), k=0,\ldots,N-1$, $X(k) = \sum_{n = 0}^{N-1} x(n)\exp(\frac{-i2\pi}{N}kn)$, By inserting 0, we have a 2N-point sequence $x'(n), n = 0, \ldots, 2N-1$, where $x'(2n) = x(n)$ and $x'(2n+1) = 0$ for $n = 0,\ldots,N-1$. Assume the DFT of $x'(n)$ is $X'(k)$. For $k<N$, considering the inserted 0,
\begin{equation}
\begin{aligned}
    X'(k) &= \sum_{n = 0}^{2N-1} x'(n)\exp(\frac{-i2\pi}{2N}kn) \\
          &= \sum_{n = 0}^{N-1} x(n)\exp(\frac{-i2\pi}{2N}k(2n))=X(k).
\end{aligned}
\end{equation}
For $k \geq N$, let $k' = k - N$, thus $k'=0,\ldots,N-1$, and,  
\begin{equation}
\begin{aligned}
    X'(k) &= \sum_{n = 0}^{N-1} x(n)\exp(\frac{-i2\pi}{2N}(k'+N)(2n)) \\
    &= \sum_{n = 0}^{N-1}(x(n)\exp(\frac{-i2\pi}{N}nk'-i2n\pi)) = X(k')
\end{aligned} 
\end{equation}
The final equality is due to the periodic property of the complex exponential function. 
Thus, there will be two copies of the previous low-resolution spectrum, one at $[0,N-1]$ and the other at $[N,2N-1]$. 
To avoid such artifacts to persist in the final output image, the high frequency component needs be removed or at least reduced. Therefore, the subsequent convolution kernels in Fig.~\ref{fig:upsampling} generally need to be low-pass filters. 
For illustration propose, in the rest of the paper, we shift the spectrum such that the low frequency components are at the center of the spectrum. 

\begin{figure}[tbp]
\begin{center}
	\includegraphics[width=0.9\linewidth]{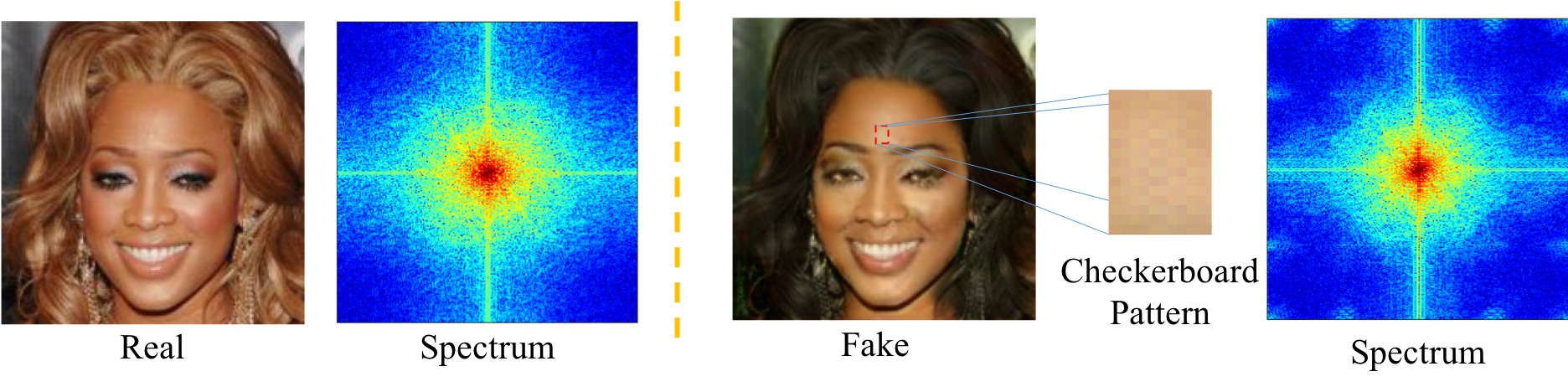}
\end{center}
\vspace{-1.5em}
\caption{The spectra of a real face image and a fake face image generated by this image. Note the checkerboard artifact in the zoomed out details.}
\label{fig:checkerboard}
\vspace{-1.5em}
\end{figure}

If the up-sampler is the transposed convolution, it's not guaranteed that the learned convolution kernel is low-pass. Thus, the checkerboard artifact can still be observed in many images. 
We show one example in Fig.~\ref{fig:checkerboard}, where the two leftmost images are one real face image and its spectrum. Images on the right are a fake face generated from the real image and its spectrum. The checkerboard artifact is highlighted in the red box. 
In the spectrum, there are bright blobs at $1/4$ and $3/4$ of the width/height, corresponding to the artifact generated by the two successive up-sampling modules in the generator.
It's the checkerboard artifact in the frequency domain.  
The nearest neighbor up-sampler uses a convolution kernel that is a fixed low-pass filter, it does eliminate the artifacts better. 
However, the artifacts are still not completely removed. 
If the low-pass filter removes too much high frequency content, the final image may become too blurry and thus easily distinguishable from the real images.  

Motivated by the theoretical properties discovered above, we propose to train GAN fake image classifiers using image spectrum as input, rather than raw pixels. 

\section{Detecting and Simulating the GAN Artifact}
\label{sec:method}
\subsection{Classifier with Spectrum Input}
\label{sec:Spectrum}

To make the classifier recognize the artifact in the frequency domain, instead of using raw image pixels, we propose to use frequency spectrum to train the classifier. 

Specifically, given an image $I$ as input, we apply the 2D DFT to each of the RGB channels and get 3 channels of frequency spectrum $F$ (the phase information is discarded). 
We compute the logarithmic spectrum $\log(F)$
and normalize the logarithmic spectrum to $[-1,1]$. 
The normalized spectrum is the input to the fake image classifier. 
The main goal of the classifier would thus be to reveal the artifacts identified in the previous section to classify an image as being generated by a GAN model or not.
The spectrum-based classifier achieves better performance than the pixel-based classifier, especially when the training data contains images from only one semantic category~(Sec.\ref{sec:exp:singlecategory}). 

To further allow us to train a classifier without fake images, in the next section, we propose \emph{AutoGAN}, which is a GAN simulator that can synthesize GAN artifacts in any image without needing to access any pre-trained GAN model. 


\subsection{AutoGAN}
\label{sec:autoGAN}
\begin{figure}[b]
\vspace{-1.5em}
\begin{center}
	\includegraphics[width=0.9\linewidth]{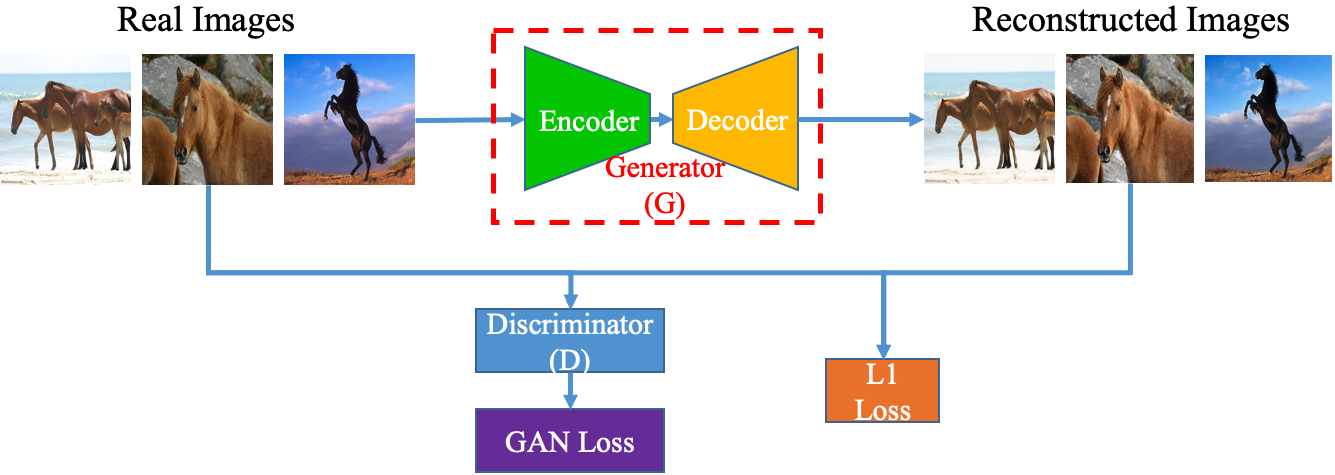}
\end{center}
\vspace{-1.0em}
\caption{The pipeline of AutoGAN.}
\label{fig:AutoGAN}
\vspace{-1.5em}
\end{figure}


We illustrate AutoGAN in Fig.~\ref{fig:AutoGAN}, which takes a real image ($I$) as input and passes it through a generator ($G$) that has a structure similar to the generator used in image generation GANs. 
The decoder contains up-sampling module such as transposed convolution or nearest neighbor interpolation. Note the only knowledge assumed available is the general architecture, but not the specifics (e.g., model weights and meta parameters) of GAN models used in fake image generation. 
Conceptually, this can be considered as a ``grey-box'' solution, compared to the ``white-box'' solution where all details of the model are known or the ``black-box'' solution where zero knowledge is available about the attack model.

The AutoGAN incorporates a discriminator ($D$) and an $\ell_1$-norm loss. 
Instead of making the distribution of the output from the generator to be similar to that of images of another semantic category, as show in general image2image translation pipeline~(Fig.~\ref{fig:pipeline}), the output of the generator is matched to the original image itself. 
Formally, assuming there is a training set $\{I_1,\ldots,I_n\}$ containing $n$ images, the final loss function $L$ can be written as, 
\begin{equation}
    L = \sum_{i=1}^{n} log(D(I_i))+log(1-D(G(I_i)))+\lambda \parallel I_i-G(I_i)\parallel_1
\end{equation}
where $D(\cdot)$ is the discriminator, $G(\cdot)$ is the generator, $G(I_i)$ is the output of $G(\cdot)$ when taking $I_i$ as input and $\lambda$ is the trade-off parameter between two different losses. 
The first two terms are similar to the GAN loss, where the discriminator wants to distinguish between the generator output and the real image, while the generator wants to fool the discriminator. 
The third term is the $\ell_1$-norm loss function to make the input and output similar in terms of $\ell_1$ distance. 

\begin{figure}[b]
\vspace{-1.5em}
\begin{center}
\includegraphics[width=1.0\linewidth]{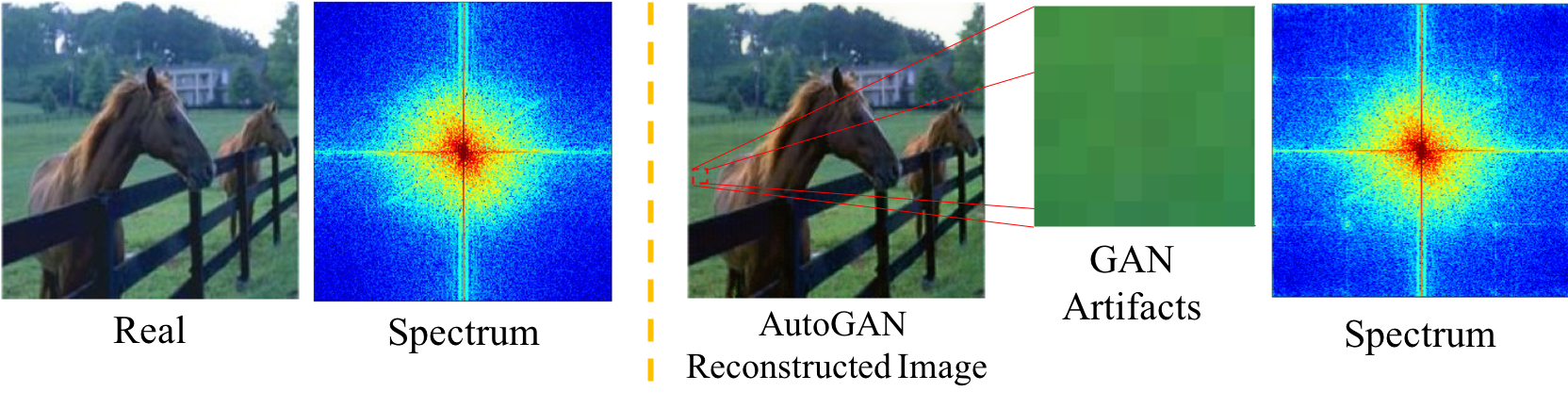}
\end{center}
\vspace{-2.0em}
\caption{The original image and the AutoGAN reconstructed image}
\label{fig:AutoGAN_image}
\vspace{-1.5em}
\end{figure}

We show one of the real image, the corresponding AutoGAN reconstructed image and their frequency spectra in Fig.~\ref{fig:AutoGAN_image}. 
Although the reconstructed image looks very similar to the real image,
there are some artifacts in the reconstructed image, especially in the frequency domain, capturing the unique GAN induced artifacts as discussed in Sec.~\ref{sec:upsampler} earlier. 
By training a classifier with the real and the reconstructed images, the classifier will focus on the artifact and thus can generalizes to other GAN fake images that have similar artifact.
The AutoGAN pipeline provides two major benefits:
\begin{itemize}
    \item It does not require any fake image in training. It only emulates and incorporates the artifacts induced by GAN pipeline into a real image. 
    \item Unlike training an image2image translation model which requires images from a pair of carefully-selected categories (such as horse and zebra), AutoGAN can take images from any semantic category as input. This greatly simplifies the data collection process.
\end{itemize}

\section{Experiment}
\subsection{Dataset}
Following~\cite{marra_detection_2018}, we conduct experiments on CycleGAN~\cite{zhu_unpaired_2017} images. 
We split the dataset based on different semantic categories. 
For example, in the horse category, there are only real horse images and fake horse images generated from zebra images. 
We use the training and test split of~\cite{zhu_unpaired_2017}.
Only images from the training set are used in training the classifier. 
And the trained classifier is tested with test images.
We also exclude the sketch and pixel-level semantic map from the dataset. 
%
There are a total number of 14 categories: Horse~(H, 2,401/260), Zebra~(Z, 2,401/260), Yosemite Summer~(S, 2,193/547), Yosemite Winter~(W, 2,193/547), Apple~(A, 2,014/514), Orange~(O, 2,014/514), Facades~(F, 800/212), CityScape\_Photo~(City, 5,950/1000), Satellite Image~(Map, 2,192/2196), Ukiyoe~(U, 2,062/1,014), Van Gogh~(V, 1,900/1,151), Cezanne~(C, 2,025/809), Monet~(M, 2,572/872) and Photo~(P, 7,359/872). 
The numbers behind each category are the numbers of training/test images in that category.

\subsection{Implementation Detail}
We use resnet34 pre-trained with ImageNet as base network and treat the GAN detection task as a binary classification problem: real vs fake.
In the training phase, we randomly crop a $224\times224$ region from the original image of size $256\times256$. 
In the test phase, the central $224\times224$ crop is used. 
The batch size is 16. The optimizer is SGD with momentum and the starting learning rate is 0.01, with a decay of $1e-2$ after each epoch. We set $\lambda=0.5$. 
The network is trained for 10 epochs. 

To train the AutoGAN model, we use the same generator and discriminator structures and the hyper-parameters detailed in \cite{zhu_unpaired_2017}. 
We only use training images from one semantic category, e.g. horse, to train the AutoGAN model. The number of training images for each category is: Horse~(1,067), Zebra~(1,334), Yosemite Summer~(1,231), Yosemite Winter~(962), Apple~(995), Orange~(1,019), Facades~(400), CityScape\_Photo~(2,975), Satellite Image~(Map, 1,096), Ukiyoe~(U, 562), Van Gogh~(400), Cezanne~(525), Monet~(M, 1,072) and Photo~(6,287). 
Our implementation is available at \url{https://github.com/ColumbiaDVMM/AutoGAN}.

\begin{table*}[htp]
\begin{center}
\begin{small}
\begin{tabular}{c|c|cccccccccccccc|c}
\hline
Training & Feature & H & Z & S & W & A & O & F & City & Map & U & V & C & M & P & Ave. \\
\hline
\multirow{4}{0.8cm}{\centering H} & Img & 99.2 & 78.5 & 96.2 & 86.3 & 78.0 & 70.6 & 75.5 & 72.2 & 55.9 & 61.5 & 95.0 & 87.0 & 87.7 & 93.6 & 81.2 \\
& Spec & 100 & 99.6 & 99.8 & 85.0 & 99.4 & 99.8 & 98.6 & 96.7 & 50.0 & 96.3 & 83.1 & 99.4 & 93.1 & 99.2 & 92.9 \\
& A Img & 91.9 & 72.7 & 87.9 & 77.3 & 83.7 & 89.1 & 52.4 & 50.4 & 57.7 & 29.0 & 61.9 & 34.9 & 36.9 & 89.0 & 65.3 \\
& A Spec & 98.1& 98.1& 99.3& 88.7& 99.6& 100& 100& 96.0& 63.5& 99.2& 86.2& 99.1& 88.1& 100& \textbf{94.0} \\
\hline
\multirow{4}{0.8cm}{\centering Z} & Img & 68.8& 96.5& 78.8& 63.4& 54.1& 50.2& 50.0& 50.0& 50.0& 39.5& 87.2& 45.4& 80.3& 87.6& 64.4 \\
& Spec & 98.1& 100& 92.5& 74.6& 97.5& 97.1& 100& 93.9& 50.0& 91.1& 53.4& 91.0& 55.5& 98.1& \textbf{85.2} \\
& A Img & 93.8& 92.7& 82.6& 84.1& 79.4& 82.1& 50.0& 50.0& 51.4& 38.5& 75.9& 49.4& 57.5& 87.0& 69.6 \\
& A Spec & 76.9& 88.8& 94.7& 52.1& 81.5& 77.6& 99.5& 80.4& 55.9& 97.2& 60.6& 97.8& 61.2& 99.0& 80.2 \\
\hline
\multirow{4}{0.8cm}{\centering S} & Img & 90.8& 87.7& 99.3& 92.0& 81.5& 80.9& 75.9& 92.4& 55.1& 92.8& 93.9& 94.4& 90.4& 93.7& 87.2 \\
& Spec & 99.2& 98.1& 100& 97.6& 82.1& 87.7& 57.1& 94.7& 50.0& 97.9& 99.4& 100& 99.4& 95.6& 89.9 \\
& A Img & 89.2& 83.1& 98.2& 98.5& 70.6& 68.9& 81.1& 63.2& 72.7& 89.2& 80.1& 93.2& 88.9& 93.3& 83.6 \\
& A Spec & 98.5& 98.8& 99.5& 89.0& 98.4& 99.0& 100& 100& 50.1& 98.7& 97.3& 99.9& 98.1& 98.1& \textbf{94.7} \\
\hline
\multirow{4}{0.8cm}{\centering W} & Img & 88.1& 83.8& 98.4& 99.1& 76.3& 74.7& 60.8& 83.6& 77.9& 87.2& 89.5& 93.8& 88.9& 94.5& 85.5 \\
& Spec & 96.5& 96.5& 100& 100& 81.1& 75.9& 91.5& 100& 50.0& 100& 100& 99.8& 99.0& 98.7& \textbf{92.1} \\
& A Img & 76.5& 78.8& 93.2& 95.4& 73.2& 71.8& 59.9& 48.6& 73.2& 92.1& 78.1& 93.2& 86.4& 83.9& 78.9 \\
& A Spec & 45.8& 57.7& 89.8& 74.8& 65.6& 74.7& 59.0& 54.4& 52.8& 96.8& 93.0& 74.9& 78.2& 96.8& 72.4 \\
\hline
\multirow{4}{0.8cm}{\centering A} & Img & 59.2& 60.4& 77.9& 66.2& 100& 82.9& 64.6& 50.1& 50.3& 35.0& 59.6& 48.0& 50.3& 87.3& \textbf{63.7} \\
& Spec & 58.8& 67.7& 57.6& 43.9& 100& 100& 94.3& 53.2& 50.0& 26.4& 34.8& 8.4& 15.1& 88.6& 57.1 \\
& A Img & 46.2& 53.8& 56.5& 43.5& 51.8& 48.2& 50.0& 50.0& 50.0& 25.9& 34.8& 7.2& 13.9& 86.1& 44.1 \\
& A Spec & 71.2& 77.7& 57.6& 45.9& 100& 100& 79.2& 54.4& 50.0& 28.3& 34.8& 9.8& 14.0& 87.7& 57.9 \\
\hline
\multirow{4}{0.8cm}{\centering O} & Img & 76.9& 70.4& 73.9& 57.0& 90.9& 98.4& 50.0& 50.3& 55.2& 51.7& 81.6& 85.4& 53.6& 89.4& \textbf{70.3} \\
& Spec & 46.5& 55.8& 56.7& 43.7& 99.6& 100& 64.6& 50.4& 50.0& 26.3& 34.8& 7.2& 13.9& 88.1& 52.7 \\
& A Img & 46.2& 53.8& 56.5& 43.5& 51.8& 48.2& 50.0& 50.0& 50.0& 25.9& 34.8& 7.2& 13.9& 86.1& 44.1 \\
& A Spec & 49.6& 63.1& 56.9& 43.3& 95.5& 97.9& 54.7& 50.6& 50.0& 26.0& 34.8& 7.3& 13.9& 88.0& 52.2 \\
\hline
\multirow{4}{0.8cm}{\centering F} & Img & 77.3& 71.9& 57.8& 47.0& 64.2& 67.9& 98.6& 86.4& 50.1& 37.1& 57.3& 84.8& 50.5& 85.0& \textbf{66.8} \\
& Spec & 87.3& 88.8& 74.8& 86.7& 75.3& 80.4& 100& 51.2& 50.6& 37.8& 50.7& 27.1& 55.5& 63.5& 66.4 \\
& A Img & 54.6& 55.0& 56.1& 45.2& 59.7& 58.4& 59.4& 57.0& 50.1& 25.9& 35.9& 32.8& 39.9& 84.7& 51.1 \\
& A Spec & 64.2& 69.2& 49.2& 43.3& 75.3& 73.0& 68.4& 50.3& 50.5& 21.7& 33.6& 8.4& 19.8& 60.2& 49.1 \\
\hline
\multirow{4}{0.8cm}{\centering City} & Img & 74.2& 65.8& 57.2& 45.3& 73.3& 74.7& 58.5& 100& 50.0& 41.9& 73.2& 92.2& 37.7& 86.0& 66.4 \\
& Spec & 93.1& 96.2& 91.0& 77.9& 63.0& 62.6& 100& 100& 50.0& 83.7& 93.5& 99.6& 94.0& 83.9& \textbf{84.9} \\
& A Img & 53.8& 46.2& 43.5& 56.5& 48.2& 51.8& 50.0& 50.0& 50.0& 74.1& 65.2& 92.8& 86.1& 13.9& 55.9 \\
& A Spec & 88.1& 95.0& 69.1& 57.0& 88.7& 87.5& 100& 87.3& 50.4& 47.5& 46.7& 70.6& 51.0& 85.2& 73.2 \\
\hline
\multirow{4}{0.8cm}{\centering Map} & Img & 71.9& 57.3& 85.6& 86.7& 60.5& 60.7& 50.0& 50.0& 93.1& 97.7& 84.5& 93.2& 87.6& 75.7& 75.3 \\
& Spec & 81.9& 65.0& 91.8& 94.9& 65.0& 72.8& 72.6& 70.9& 100& 97.1& 92.1& 95.7& 90.0& 86.9& \textbf{84.1} \\
& A Img & 55.8& 49.2& 58.3& 68.9& 48.2& 51.9& 50.5& 75.3& 69.5& 75.0& 66.1& 92.8& 86.2& 32.8& 62.9 \\
& A Spec & 79.6& 70.4& 91.6& 96.2& 59.5& 51.8& 84.4& 79.5& 92.0& 94.0& 83.9& 94.6& 91.5& 83.7& 82.3 \\
\hline
\multirow{4}{0.8cm}{\centering U} & Img & 53.8& 46.2& 43.5& 56.5& 48.2& 51.8& 50.0& 50.0& 50.0& 74.1& 65.2& 92.8& 86.1& 13.9& 55.9 \\
& Spec & 98.1& 95.8& 99.6& 96.9& 88.5& 87.0& 98.6& 100& 50.0& 100& 99.7& 100& 99.9& 99.0& \textbf{93.8} \\
& A Img & 60.0& 56.2& 69.1& 57.2& 57.2& 59.1& 50.9& 58.1& 54.1& 23.8& 68.7& 58.7& 71.7& 85.7& 59.3 \\
& A Spec & 84.2& 87.7& 58.9& 45.7& 89.3& 85.0& 81.6& 64.8& 50.0& 29.6& 35.1& 32.0& 19.6& 88.2& 60.8 \\
\hline
\multirow{4}{0.8cm}{\centering V} & Img & 53.8& 46.2& 43.5& 56.5& 48.2& 51.8& 50.0& 50.0& 50.0& 74.1& 65.2& 92.8& 86.1& 13.9& 55.9 \\
& Spec & 95.8& 96.9& 99.3& 96.5& 75.9& 70.6& 98.1& 99.7& 50.0& 99.1& 100& 99.9& 99.5& 97.5& \textbf{91.3} \\
& A Img & 53.8& 46.2& 43.5& 56.5& 48.2& 51.8& 50.0& 50.0& 50.0& 74.1& 65.2& 92.8& 86.1& 13.9& 55.9 \\
& A Spec & 76.9& 77.3& 58.0& 46.3& 85.4& 84.0& 82.1& 78.3& 50.0& 28.2& 35.9& 34.5& 18.2& 84.6& 60.0 \\
\hline
\multirow{4}{0.8cm}{\centering C} & Img & 53.8& 46.2& 43.5& 56.5& 48.2& 51.8& 50.0& 50.0& 50.0& 74.1& 65.2& 92.8& 86.1& 13.9& 55.9 \\
& Spec & 95.8& 95.0& 75.3& 64.7& 89.7& 91.6& 99.5& 99.5& 50.2& 80.3& 73.9& 95.9& 61.8& 91.4& \textbf{83.2} \\
& A Img & 54.6& 56.5& 58.9& 45.0& 73.9& 74.3& 50.0& 50.0& 50.0& 26.2& 37.1& 33.9& 34.9& 85.8& 52.2 \\
& A Spec & 71.5& 83.1& 58.3& 47.3& 94.4& 90.3& 95.3& 58.3& 50.3& 27.8& 34.8& 11.2& 15.6& 85.6& 58.8\\
\hline
\multirow{4}{0.8cm}{\centering M} & Img & 86.2& 80.0& 93.1& 89.9& 60.1& 58.9& 57.5& 69.6& 72.4& 60.9& 87.7& 91.1& 99.7& 77.3& 77.5 \\
& Spec & 99.6& 97.7& 99.3& 97.4& 90.5& 92.4& 99.1& 100& 50.1& 99.9& 100& 100& 100& 99.2& \textbf{94.7} \\
& A Img & 48.8& 54.2& 56.3& 43.7& 53.5& 54.1& 50.0& 50.1& 50.0& 25.9& 35.4& 10.0& 15.6& 86.1& 45.3 \\
& A Spec & 67.7& 83.5& 58.7& 46.6& 98.6& 100& 74.5& 53.4& 50.3& 29.3& 34.8& 11.7& 14.7& 87.7& 58.0 \\
\hline
\multirow{4}{0.8cm}{\centering Photo} & Img & 85.4& 86.2& 97.4& 95.6& 68.1& 62.5& 88.2& 93.8& 69.4& 97.2& 77.2& 92.6& 82.6& 99.5& 85.4 \\
& Spec & 96.2& 94.6& 96.5& 62.9& 98.1& 99.6& 80.7& 99.3& 51.9& 98.1& 77.7& 98.4& 86.5& 100& 88.6 \\
& A Img & 89.2& 86.9& 99.1& 99.5& 74.3& 72.6& 59.0& 65.4& 94.9& 97.5& 83.3& 93.3& 88.6& 94.7& 85.6 \\
& A Spec & 100& 99.6& 100& 98.9& 96.3& 97.5& 100& 100& 53.6& 100& 99.9& 100& 100& 99.8& \textbf{96.1} \\
\hline
\multirow{2}{0.8cm}{COCO} & A Img & 77.3 & 78.8 & 58.7 & 75.3 & 64.8 & 69.5 & 100 & 99.9 & 73.4 & 88.9 & 89.0 & 97.4 & 91.5 & 37.7 & 78.7\\
& A Spec & 90.4 & 90.4 & 83.7 & 85.2 & 94.0 & 93.8 & 99.5 & 100 & 88.9 & 97.8 & 91.6 & 98.6 & 96.4 & 81.8 & \textbf{92.3}\\
\hline
\end{tabular}%
\end{small}
\end{center}
\caption{Test accuracy over all semantic categories using one category for training.}
\label{table:single_category}
\vspace{-1.5em}
\end{table*}

\begin{table*}[htp]
\begin{center}
\begin{small}
\begin{tabular}{c|cccccccccc|c}
\hline
Method & H-Z & S-W & A-O & F & City & Map & U & V & C & M & Ave. \\
\hline
Cozzalino2017~\cite{marra_detection_2018} & 99.9 & 100 & 61.2 & 99.9 & 97.3 & 99.6 & 100 & 99.9 & 100 & 99.2 & 95.7 \\
DenseNet~\cite{marra_detection_2018} & 79.1 & 95.8 & 67.7 & 99.0 & 93.8 & 78.3 & 99.5 & 97.7 & 99.9 & 89.8 & 90.1 \\
XceptionNet~\cite{marra_detection_2018} & 95.9 & 99.2 & 76.7 & 100 & 98.6 & 76.8 & 100 & 99.9 & 100 & 95.1 & 94.2 \\
Nataraj2019~\cite{nataraj_detecting_2019} & 99.8 & 99.8 & 99.7 & 92.0 & 80.6 & 97.5 & 99.6 & 100 & 99.6 & 99.2 & 96.8 \\
Img & 94.7 & 81.9 & 99.5 & 98.6 & 94.7 & 58.8 & 99.4 & 99.9 & 99.9 & 92.5 & 92.0 \\
Spec & 99.8 & 99.8 & 99.8 & 100 & 99.9 & 60.7 & 100 & 98.6 & 100 & 100 & 95.9 \\
A-Img &  87.2 & 93.3 & 98.4 & 92.8 & 50.7 & 52.0 & 61.5 & 93.2 & 71.5 & 86.7 & 78.7 \\
A-Spec & 98.4 & 99.9 & 98.3 & 100 & 100 & 78.6 & 99.9 & 97.5 & 99.2 & 99.7 & \textbf{97.2} \\
\hline
\end{tabular}%
\end{small}
\end{center}
\caption{Test accuracy using the leave-one-out setting in \cite{marra_detection_2018}. A-Img and A-Spec are models based on AutoGAN.}
\label{table:leave_one_out}
\vspace{-2.0em}
\end{table*}

\subsection{Training with a Single Semantic Category}
\label{sec:exp:singlecategory}
We first show the performance of the classifiers trained with images from one single category. 
Four different types of classifiers are trained for comparison:
\begin{itemize}
    \item \textbf{Img}: Learned with real images and fake images generated by cycleGAN (for example, real horse images and fake horse images generated from zebra images);
    \item \textbf{Spec}: The training data is the same as \textbf{Img}, the classifier is trained with the spectrum input;
    \item \textbf{A-Img}: Learned with real image and fake image generated by AutoGAN (for example, real horse images and reconstructed horse images from AutoGAN);
    \item \textbf{A-Spec}: The training data is the same as \textbf{A-Img}, the classifier is trained with the spectrum input.
\end{itemize}

\begin{figure*}
\begin{center}
\subfigure[Horse]{
	\includegraphics[width=0.40\linewidth]{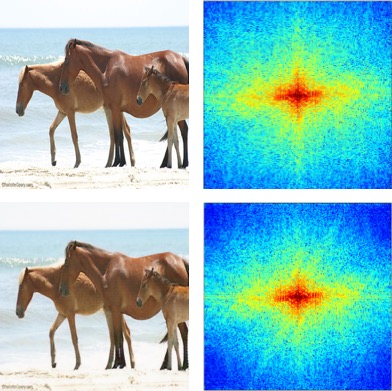}
	\label{fig:Horse}
    }
\hspace{3em}
\subfigure[Zebra]{
	\includegraphics[width=0.40\linewidth]{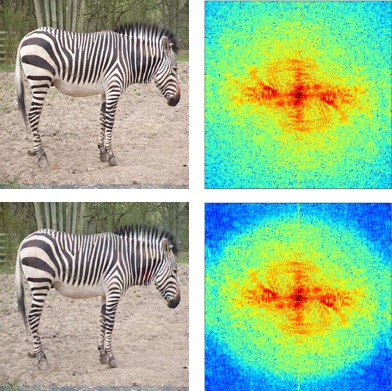}
	\label{Zebra}
    } \\ 
\subfigure[Summer]{
	\includegraphics[width=0.40\linewidth]{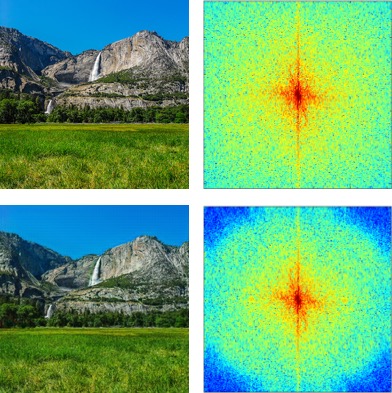}
	\label{Summer}
    } 
\hspace{3em}
\subfigure[Winter]{
	\includegraphics[width=0.40\linewidth]{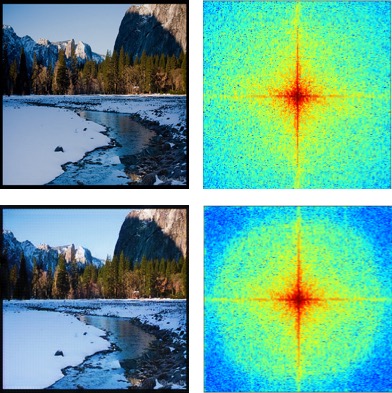}
	\label{Winter}
    } \\ 
\subfigure[Apple]{
	\includegraphics[width=0.40\linewidth]{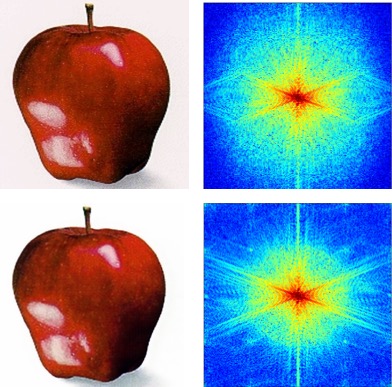}
	\label{Apple}
    } 
\hspace{3em}
\subfigure[Orange]{
	\includegraphics[width=0.40\linewidth]{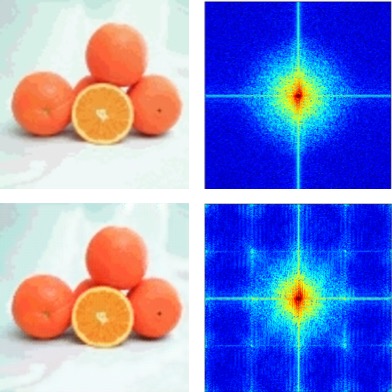}
	\label{Orange}
    } \\ 
\vspace{-1.0em}
\caption{Sample images and spectra for each category. For each sub image, the images in the first row are the original image and its spectrum, while the images in the second row are the  AutoGAN image and its spectrum.}
\label{fig:image}
\vspace{-1.0em}
\end{center}
\end{figure*}

\begin{figure*}
\begin{center}
\subfigure[Facades]{
	\includegraphics[width=0.40\linewidth]{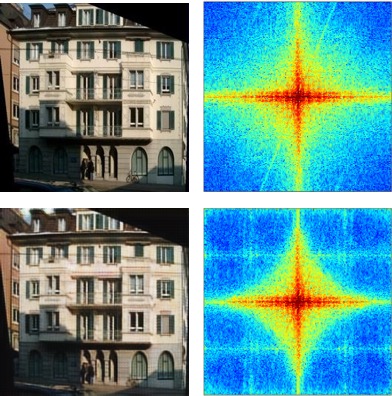}
	\label{Facades}
    }
\hspace{3em}
\subfigure[Cityscape Photo]{
	\includegraphics[width=0.40\linewidth]{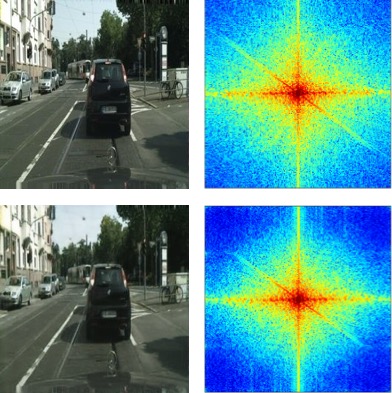}
	\label{City}
    } \\ 
\subfigure[Satellite Image]{
	\includegraphics[width=0.40\linewidth]{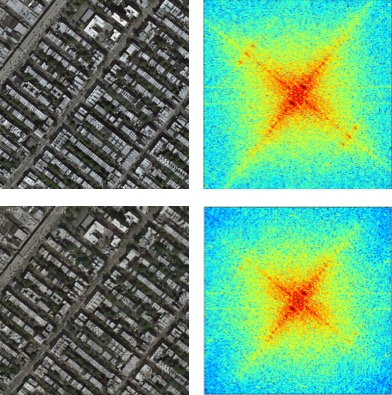}
	\label{Map}
    } 
\hspace{3em}
\subfigure[Ukiyoe]{
	\includegraphics[width=0.40\linewidth]{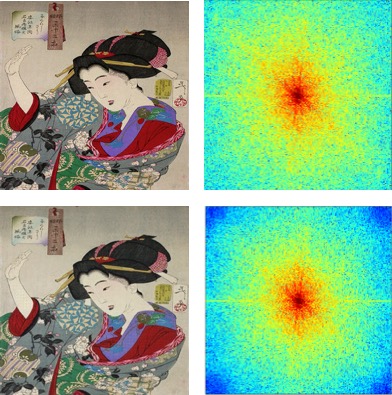}
	\label{Ukiyoe}
    } \\ 
\subfigure[Van Gogh]{
	\includegraphics[width=0.40\linewidth]{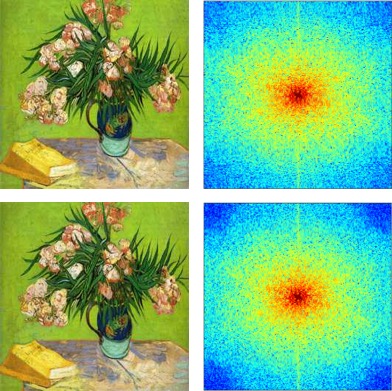}
	\label{vangogh}
    } 
\hspace{3em}
\subfigure[Cezanne]{
	\includegraphics[width=0.40\linewidth]{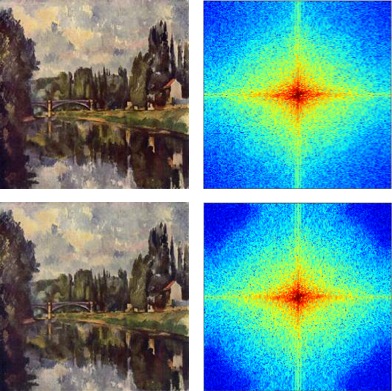}
	\label{Cezanne}
    } \\ 
\caption{Sample images and spectra for each category (continued).}
\label{fig:image_2}
\vspace{-1.0em}
\end{center}
\end{figure*}

\begin{figure*}
\begin{center}
\subfigure[Monet]{
	\includegraphics[width=0.40\linewidth]{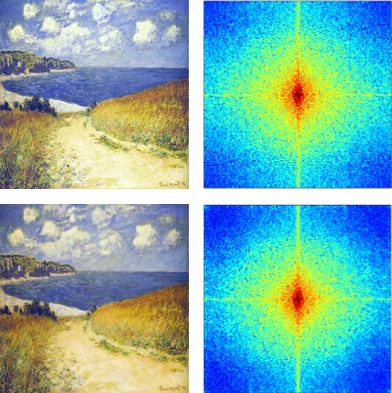}
	\label{Monet}
    }
\hspace{3em}
\subfigure[Photo]{
	\includegraphics[width=0.40\linewidth]{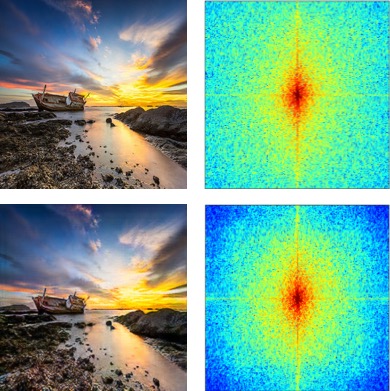}
	\label{Photo}
    } 
\caption{Sample images and spectra for each category (continued).}
\label{fig:image_3}
\vspace{-1.0em}
\end{center}
\end{figure*}

All 14 categories from CycleGAN are used for training. 
The classifiers are evaluated on all the categories and the test accuracy is reported in Table~\ref{table:single_category}. Sample images and corresponding spectra are shown from Fig.\ref{fig:image} - Fig.\ref{fig:image_3}.
When trained with cycleGAN images (Img and Spec), if the training category and test category are the same, e.g. training and testing with horse images, the accuracy is close to perfect. 
However the classifier trained with cycleGAN image (Img) struggles to generalize well to other categories.
The spectrum-based classifier (Spec) greatly improves the generalization ability, 
indicating the spectrum-based classifier is able to discover some common artifacts induced by a GAN pipeline. 
The exceptions are Apple, Orange and Facades, whose original images have been heavily compressed and don't show a lot of high frequency component compared to others. When used as test category, Map seems to be an outlier, since all categories except itself can't achieve promising performance on it. The reason is that the satellite image (Fig. \ref{Map}) is very different from other images since it's taken from a very high altitude, thus all building are small and regular. The difference is quite obvious in the frequency domain, since it shows tilted pattern compared to all others. Overall, the training image for the spectrum detector needs to have good coverage and be representative.

When trained with AutoGAN generated images (A-Img), the performance of the classifier is inferior. 
The reason is that the images generated by AutoGAN and cycleGAN are quite different, especially in terms of image quality. 
Training with the image pixels as input may suffer from this domain gap. 
When trained with the spectrum of the AutoGAN generated image (A-Spec), the performance of the classifier is quite promising with some of the categories (Horse, Summer and Photo). 
It shows that the proposed GAN simulator is able to simulate the artifacts induced by GAN pipeline.
\emph{However, the training category needs to be carefully selected. }
It requires the spectrum of the training images to be diverse enough for generalization. 
For example, due to the stride in zebra images as well as the snow in winter images, the zebra images and the winter images may not have good coverage over the frequency domain. 
Another interesting finding is that all the artwork categories~(Ukiyoe, Vangogh, Cezanne and Monet), although their performances with spectrum classifier are satisfactory, their performances with AutoGAN spectrum classifier are not promising. There may have two reasons: 1) the number of training images for AutoGAN spectrum classifier is smaller than that used in the spectrum classifier training; 2) When training the spectrum classifiers, the fake images generated from real photo category are available. Those images still contain information form real photos, allowing the classifier to see diverse spectrum.

One solution to get sufficient spectrum coverage is to use a diverse image dataset such as ImageNet or MSCOCO to train the AutoGAN model. 
We randomly select 4,000 images from MSCOCO to train an AutoGAN model. 
The results of the classifier trained with real and reconstructed MSCOCO images from AutoGAN are shown in the ``COCO'' category of Table~\ref{table:single_category}. 
Although the spectrum-based classifier trained with COCO images has never seen any cycleGAN images during training, it still works reasonably well. 
This again proves that the AutoGAN can capture the general GAN artifacts.

\subsection{Leave-One-Out Setting}
\label{sec:exp:leaveoneout}
To compare with the state-of-the-art methods, we follow the experimental setting detailed in \cite{marra_detection_2018} by splitting all the images into 10 folds and perform a leave-one-out test. 
Following~\cite{marra_detection_2018}, we use a pre-trained DenseNet121 model as base network. 
The results are shown in Table~\ref{table:leave_one_out}. 
The spectrum-based classifiers trained with CycleGAN images (Spec) and AutoGAN images (A-Spec) are competitive with the state-of-the-art methods. 
Considering A-Spec has never seen any fake image generated by any pre-trained CycleGAN model, it shows the effectiveness of the GAN simulator.

\subsection{Effect of Different Frequency Bands}
\begin{figure}[htp]
\begin{center}
	\includegraphics[width=0.8\linewidth]{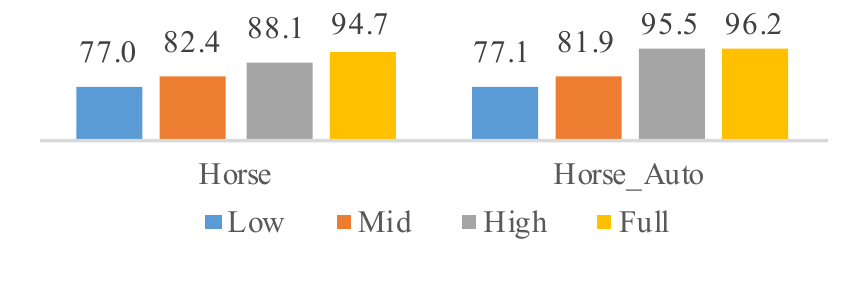}
\end{center}
\vspace{-2.5em}
\caption{Performance of models trained with different frequency bands.}
\label{fig:freq}
\vspace{-1.0em}
\end{figure}

To show which part of the spectrum affects the performance of the spectrum-based classifier, we split the full spectrum into 3 parts: low-frequency, middle-frequency and high-frequency, such that numbers of the data points in each bands are about the same. The points outside the selected band are set to 0.
We train 6 different classifiers using 3 different frequency bands and horse images from CycleGAN and AutoGAN respectively.
The performances are shown in Fig.~\ref{fig:freq}. Since the up-sampling artifacts appear in mid and high frequency bands, the classifiers trained with mid and high frequency bands show better performances. 

\subsection{Robustness to Post-Processing}
We show the robustness of the proposed method with two different post-processing methods: JPEG compression and image resize. 
For JPEG compression, we randomly select one JPEG quality factor from [100, 90, 70, 50] and apply it to each of the fake image. For image resize, we randomly select one image size from [256, 200, 150, 128].  

We show the performances of 2 types of models in Table~\ref{table:attack}, 
1) trained with images without post-processing and tested with images subject to post-processing (Mismatched), and 
2) trained and tested with images subject to post-processing (Retrained). 
Performances of models trained and tested with images without any post-processing are also given (Original). 
We use horse images from CycleGAN and AutoGAN for training respectively. 
Since both the JPEG compression and image resize destroy the up-sampling artifact, the model trained with images without post-processing does not generalize to the post-processed images, also as reported in~\cite{nataraj_detecting_2019}.  
Training a new model with post-processed images improves the performance on post-processed images. Note we do not assume the retained model has information about the specific resize factor or JPEG quality factor used in each test image. Each training image used in the retrained model uses randomly selected factors.
\begin{table}[htp]
\begin{center}
\begin{small}
\begin{tabular}{c|c|c|cc}
\hline
 Attack & Training & Original & Mismatched & Retrained \\
\hline
\multirow{2}{1.0cm}{JPEG}& Horse Spec  & 92.9 & 48.1 & 89.3 \\
& Horse A-Spec & 94.0 & 49.6 & 85.5 \\
\hline
\multirow{2}{1.0cm}{Resize}& Horse Spec  & 92.9 & 74.8 & 97.3 \\
& Horse A-Spec & 94.0 & 67.0 & 98.1 \\
\hline
\end{tabular}%
\end{small}
\end{center}
\caption{Test accuracy against different post-processing methods.}
\label{table:attack}
\vspace{-1.0em}
\end{table}

\subsection{Generalization Ability}
\subsubsection{Generalization to Different Up-samplers}
As the artifacts are induced by the up-sampler in the GAN pipeline, we are interested in assessing the performance on images generated by GAN with different up-samplers. 
We tested 2 different up-samplers, transposed convolution (Trans.)~\cite{zhu_unpaired_2017} and nearest neighbor interpolation (NN)~\cite{karras_progressive_2017}. 
We change the Trans. up-sampler in CycleGAN/AutoGAN to the NN up-sampler used in~\cite{karras_progressive_2017}. 
For each up-sampler, we train 6 image2image translation CycleGAN models, transferring between horse$\leftrightarrow$zebra, summer$\leftrightarrow$winter and apple$\leftrightarrow$orange. 
We also train 6 AutoGAN models with different up-samplers. The results are shown in Table~\ref{table:different_upsampler}.

If the classifier is trained and tested with images generated by the same up-sampler, the performance is very good.
However, if there is mismatch between the up-samplers, the performance drops noticeably, especially when trained with Trans. and tested with NN.  
The reasons are 1) the artifacts induced by different up-samplers are different; 2) NN generated images have less artifacts~\cite{odena_deconvolution_2016}. 
To address this issue, we can train the classifier using data generated from both up-samplers. This approach (Comb.) works well and the model achieves excellent performance for both up-samplers (Table~\ref{table:different_upsampler}).

\begin{table}[htp]
\begin{center}
\begin{small}
\begin{tabular}{c|ccc|ccc}
\hline
\multirow{2}{0.8cm}{Test}& \multicolumn{3}{c|}{CycleGAN} & \multicolumn{3}{c}{AutoGAN}\\
 & Trans. & NN & Comb. & Trans. & NN & Comb. \\
\hline
Trans. & 93.1 & 89.4 & 93.4 & 98.9 & 79.2 & 97.4 \\
NN & 54.0 & 97.5 & 96.5 & 70.5 & 93.9 & 95.4 \\
\hline
\end{tabular}%
\end{small}
\end{center}
\caption{Test accuracy of models trained with different up-samplers.}
\label{table:different_upsampler}
\vspace{-1.0em}
\end{table}

\subsubsection{Generalization to Different Models}
We further test the generalization ability of the GAN classifier over images generated with different GAN models. 
StarGAN~\cite{choi_stargan:_2017} and GauGAN~\cite{park_semantic_2019} are chosen as the test models. We tested 4 classifiers, image-based and spectrum-based classifiers trained with either CycleGAN or AutoGAN images. 
The results are shown in Table~\ref{table:different_models}. 
The cycleGAN image classifier fails at StarGAN images, while the spectrum-based classifiers work well. 
This again shows the generalization ability of the spectrum-based classifier. 
Note that StarGAN and CycleGAN have similar generators (with 2 transposed convolution up-samplers).
However, all classifiers fail at GauGAN images, since the generator structure of GauGAN~(5 nearest neighbor up-samplers) is drastically different from the CycleGAN structure.
%
\begin{table}[htp]
\begin{center}
\begin{small}
\begin{tabular}{c|cccc}
\hline
\multirow{2}{0.8cm}{Test}& \multicolumn{4}{c}{Trained with CycleGAN Images}\\
 & Img & Spec & A-Img & A-Spec \\
\hline
StarGAN & 65.1 & 100 & 92.5 & 98.7\\
GauGAN & 59.4 & 50.3 & 56.6 & 50.0\\
\hline
\end{tabular}%
\end{small}
\end{center}
\caption{Test accuracy tested with unseen models}
\label{table:different_models}
\vspace{-1.0em}
\end{table}

\section{Conclusion}
We study the artifacts induced by the up-sampler of the GAN pipelines in the frequency domain, in order to develop robust GAN fake image classifiers.
To detect such artifacts, we propose to use the frequency spectrum instead of image pixels as input for classifier training. 
It greatly improves the generalization ability of the classifier. 
We further propose AutoGAN which simulates the common GAN pipeline and synthesizes GAN artifacts in real images. 
The proposed AutoGAN allows us to train a GAN fake image classifier without needing fake images as training data or specific GAN models used for generating fake images. 
The AutoGAN spectrum-based classifier generalizes well to fake images generated by GANs with similar structures. Our future work includes extension of the proposed GAN simulator to study other processing modules besides the up-sampling module.

\section*{Acknowledgment}
This material is based upon work supported by the United States Air Force Research Laboratory (AFRL) and the Defense Advanced Research Projects Agency (DARPA) under Contract No.~FA8750-16-C-0166. Any opinions, findings and conclusions or recommendations expressed in this material are solely the responsibility of the authors and does not necessarily represent the official views of AFRL, DARPA, or the U.S. Government.

\bibliographystyle{IEEEtran}
\bibliography{IEEEexample}

\begin{thebibliography}{10}
\providecommand{\url}[1]{#1}
\csname url@samestyle\endcsname
\providecommand{\newblock}{\relax}
\providecommand{\bibinfo}[2]{#2}
\providecommand{\BIBentrySTDinterwordspacing}{\spaceskip=0pt\relax}
\providecommand{\BIBentryALTinterwordstretchfactor}{4}
\providecommand{\BIBentryALTinterwordspacing}{\spaceskip=\fontdimen2\font plus
\BIBentryALTinterwordstretchfactor\fontdimen3\font minus
  \fontdimen4\font\relax}
\providecommand{\BIBforeignlanguage}[2]{{%
\expandafter\ifx\csname l@#1\endcsname\relax
\typeout{** WARNING: IEEEtran.bst: No hyphenation pattern has been}%
\typeout{** loaded for the language `#1'. Using the pattern for}%
\typeout{** the default language instead.}%
\else
\language=\csname l@#1\endcsname
\fi
#2}}
\providecommand{\BIBdecl}{\relax}
\BIBdecl

\bibitem{goodfellow2014generative}
I.~Goodfellow, J.~Pouget-Abadie, M.~Mirza, B.~Xu, D.~Warde-Farley, S.~Ozair,
  A.~Courville, and Y.~Bengio, ``Generative adversarial nets,'' in \emph{NIPS},
  2014.

\bibitem{karras_style-based_2018}
T.~Karras, S.~Laine, and T.~Aila, ``A {Style}-{Based} {Generator}
  {Architecture} for {Generative} {Adversarial} {Networks},'' \emph{CVPR},
  2019.

\bibitem{brock_large_2018}
A.~Brock, J.~Donahue, and K.~Simonyan, ``Large {Scale} {GAN} {Training} for
  {High} {Fidelity} {Natural} {Image} {Synthesis},'' \emph{ICLR}, 2019.

\bibitem{marra_detection_2018}
F.~Marra, D.~Gragnaniello, D.~Cozzolino, and L.~Verdoliva, ``Detection of
  {GAN}-{Generated} {Fake} {Images} over {Social} {Networks},'' in
  \emph{{MIPR}}, 2018.

\bibitem{yu_attributing_2018}
N.~Yu, L.~Davis, and M.~Fritz, ``Attributing {Fake} {Images} to {GANs}:
  {Analyzing} {Fingerprints} in {Generated} {Images},'' \emph{arXiv:1811.08180
  [cs]}, 2018.

\bibitem{zhu_unpaired_2017}
J.-Y. Zhu, T.~Park, P.~Isola, and A.~A. Efros, ``Unpaired {Image}-to-{Image}
  {Translation} using {Cycle}-{Consistent} {Adversarial} {Networks},''
  \emph{ICCV}, 2017.

\bibitem{choi_stargan:_2017}
Y.~Choi, M.~Choi, M.~Kim, J.-W. Ha, S.~Kim, and J.~Choo, ``{StarGAN}: {Unified}
  {Generative} {Adversarial} {Networks} for {Multi}-{Domain} {Image}-to-{Image}
  {Translation},'' \emph{CVPR}, 2018.

\bibitem{odena_deconvolution_2016}
A.~Odena, V.~Dumoulin, and C.~Olah, ``Deconvolution and {Checkerboard}
  {Artifacts},'' \emph{Distill}, 2016.

\bibitem{karras_progressive_2017}
T.~Karras, T.~Aila, S.~Laine, and J.~Lehtinen, ``Progressive {Growing} of
  {GANs} for {Improved} {Quality}, {Stability}, and {Variation},'' \emph{ICLR},
  2018.

\bibitem{park_semantic_2019}
T.~Park, M.-Y. Liu, T.-C. Wang, and J.-Y. Zhu, ``Semantic {Image} {Synthesis}
  with {Spatially}-{Adaptive} {Normalization},'' \emph{CVPR}, 2019.

\bibitem{isola2017image}
P.~Isola, J.-Y. Zhu, T.~Zhou, and A.~A. Efros, ``Image-to-image translation
  with conditional adversarial networks,'' in \emph{CVPR}, 2017.

\bibitem{li_ictu_2018}
Y.~Li, M.-C. Chang, and S.~Lyu, ``In {Ictu} {Oculi}: {Exposing} {AI}
  {Generated} {Fake} {Face} {Videos} by {Detecting} {Eye} {Blinking},''
  \emph{arXiv:1806.02877 [cs]}, 2018.

\bibitem{li_exposing_2018}
Y.~Li and S.~Lyu, ``Exposing {DeepFake} {Videos} {By} {Detecting} {Face}
  {Warping} {Artifacts},'' \emph{arXiv:1811.00656 [cs]}, 2018.

\bibitem{agarwal_protecting_nodate}
S.~Agarwal, H.~Farid, Y.~Gu, M.~He, K.~Nagano, and H.~Li, ``Protecting {World}
  {Leaders} {Against} {Deep} {Fakes},'' \emph{CVPRW}, 2019.

\bibitem{nataraj_detecting_2019}
L.~Nataraj, T.~M. Mohammed, B.~S. Manjunath, S.~Chandrasekaran, A.~Flenner,
  J.~H. Bappy, and A.~K. Roy-Chowdhury, ``Detecting {GAN} generated {Fake}
  {Images} using {Co}-occurrence {Matrices},'' \emph{arXiv:1903.06836 [cs]},
  2019.

\bibitem{mccloskey_detecting_2018}
S.~McCloskey and M.~Albright, ``Detecting {GAN}-generated {Imagery} using
  {Color} {Cues},'' \emph{arXiv:1812.08247 [cs]}, 2018.

\bibitem{marra2019gans}
F.~Marra, D.~Gragnaniello, L.~Verdoliva, and G.~Poggi, ``Do gans leave
  artificial fingerprints?'' in \emph{MIPR}, 2019.

\bibitem{albright2019source}
M.~Albright and S.~McCloskey, ``Source generator attribution via inversion,''
  \emph{CVPRW}, 2019.

\bibitem{radford_unsupervised_2015}
A.~Radford, L.~Metz, and S.~Chintala, ``Unsupervised {Representation}
  {Learning} with {Deep} {Convolutional} {Generative} {Adversarial}
  {Networks},'' \emph{arXiv:1511.06434 [cs]}, 2015.

\end{thebibliography}
\end{document}